\documentclass{article} 
\usepackage{iclr2016_workshop,times}
\usepackage{hyperref}
\usepackage{url}
\usepackage{amssymb}
\usepackage{amsmath}
\usepackage{booktabs}
\usepackage{graphicx}

\title{Feed-Forward Networks with Attention Can Solve Some Long-Term Memory Problems}

\author{Colin Raffel\\
LabROSA, Columbia University\\
\texttt{craffel@gmail.com}
\And
Daniel P.~W.~Ellis\\
LabROSA, Columbia University\\
\texttt{dpwe@ee.columbia.edu}
}

\begin{document}

\maketitle

\begin{abstract}
We propose a simplified model of attention which is applicable to feed-forward neural networks and demonstrate that the resulting model can solve the synthetic ``addition'' and ``multiplication'' long-term memory problems for sequence lengths which are both longer and more widely varying than the best published results for these tasks.
\end{abstract}

\section{Models for Sequential Data}

Many problems in machine learning are best formulated using sequential data and appropriate models for these tasks must be able to capture temporal dependencies in sequences, potentially of arbitrary length.
One such class of models are recurrent neural networks (RNNs), which can be considered a learnable function $f$ whose output $h_t = f(x_t, h_{t - 1})$ at time $t$ depends on input $x_t$ and the model's previous state $h_{t - 1}$.
Training of RNNs with backpropagation through time \citep{werbos1990backpropagation} is hindered by the vanishing and exploding gradient problem \citep{pascanu2012difficulty,hochreiter1997long,bengio1994learning}, and as a result RNNs are in practice typically only applied in tasks where sequential dependencies span at most hundreds of time steps.
Very long sequences can also make training computationally inefficient due to the fact that RNNs must be evaluated sequentially and cannot be fully parallelized.

\subsection{Attention}

A recently proposed method for easier modeling of long-term dependencies is ``attention''.
Attention mechanisms allow for a more direct dependence between the state of the model at different points in time.
Following the definition from \citep{bahdanau2014neural}, given a model which produces a hidden state $h_t$ at each time step, attention-based models compute a ``context'' vector $c_t$ as the weighted mean of the state sequence $h$ by
$$
c_t = \sum_{j = 1}^T \alpha_{tj} h_j
$$
where $T$ is the total number of time steps in the input sequence and $\alpha_{tj}$ is a weight computed at each time step $t$ for each state $h_j$.
These context vectors are then used to compute a new state sequence $s$, where $s_t$ depends on $s_{t - 1}$, $c_t$ and the model's output at $t - 1$.
The weightings $\alpha_{tj}$ are then computed by
\begin{equation*}
e_{tj} = a(s_{t - 1}, h_j), \alpha_{tj} = \frac{\exp(e_{tj})}{\sum_{k = 1}^T \exp(e_{tk})}
\end{equation*}
where $a$ is a learned function which can be thought of as computing a scalar importance value for $h_j$ given the value of $h_j$ and the previous state $s_{t - 1}$.
This formulation allows the new state sequence $s$ to have more direct access to the entire state sequence $h$.
Attention-based RNNs have proven effective in a variety of sequence transduction tasks, including machine translation \citep{bahdanau2014neural}, image captioning \citep{xu2015show}, and speech recognition \citep{chan2015listen,bahdanau2015end}.
Attention can be seen as analogous to the ``soft addressing'' mechanisms of the recently proposed Neural Turing Machine \citep{graves2014neural} and End-To-End Memory Network \citep{sukhbaatar2015end} models.

\subsection{Feed-Forward Attention}

\begin{figure}
  \centering
  \includegraphics[width=.8\textwidth]{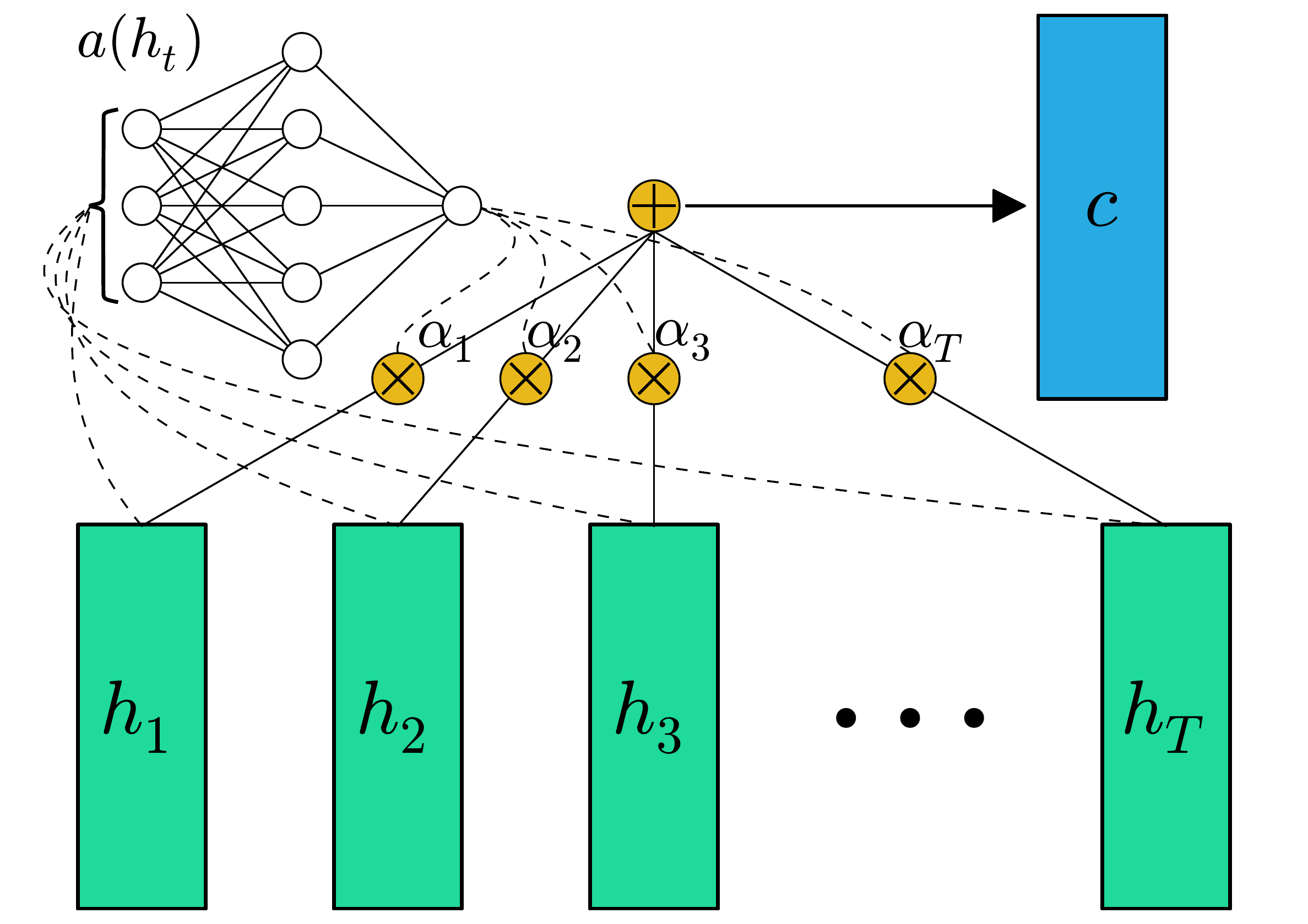}
  \caption{Schematic of our proposed ``feed-forward'' attention mechanism (cf.\ \citep{cho2015introduction} Figure 1).  Vectors in the hidden state sequence $h_t$ are fed into the learnable function $a(h_t)$ to produce a probability vector $\alpha$.  The vector $c$ is computed as a weighted average of $h_t$, with weighting given by $\alpha$.}
  \label{fig:schematic}
\end{figure}

A straightforward simplification to the attention mechanism described above which would allow it to be used to produce a single vector $c$ from an entire sequence could be formulated as follows:
\begin{equation}
\label{eq:ffattention}
e_t = a(h_t), \alpha_t = \frac{\exp(e_t)}{\sum_{k = 1}^T \exp(e_k)}, c = \sum_{t = 1}^T \alpha_t h_t
\end{equation}
As before, $a$ is a learnable function, but it now only depends on $h_t$.
In this formulation, attention can be seen as producing a fixed-length embedding $c$ of the input sequence by computing an adaptive weighted average of the state sequence $h$.
A schematic of this form of attention is shown in Figure \ref{fig:schematic}.
\cite{sonderby2015convolutional} compared the effectiveness of a standard recurrent network to a recurrent network augmented with this simplified version of attention on the task of protein sequence analysis.

A consequence of using an attention mechanism is the ability to integrate information over time.
It follows that by using this simplified form of attention, a model could handle variable-length sequences even if the calculation of $h_t$ was feed-forward, i.e.\ $h_t = f(x_t)$.
Using a feed-forward $f$ could also result in large efficiency gains as the computation could be completely parallelized.
We investigate the capabilities of this ``feed-forward attention'' model in Section \ref{sec:experiments}.

We note here that feed-forward models without attention can be used for sequential data when the sequence length $T$ is fixed, but when $T$ varies across sequences, some form of temporal integration is necessary.
An obvious straightforward choice, which can be seen as an extreme oversimplification of attention, would be to compute $c$ as the unweighted average of the state sequence $h_t$, i.e.
\begin{equation}
\label{eq:unweighted}
c = \frac{1}{T}\sum_{t = 1}^T h_t
\end{equation}
This form of integration has been used to collapse the temporal dimension of audio \citep{dieleman2014recommending} and text document \citep{lei2015molding} sequences.
We will also explore the effectiveness of this approach.

\section{Toy Long-Term Memory Problems}
\label{sec:experiments}

A common way to measure the long-term memory capabilities of a given model is to test it on the synthetic problems originally proposed by \cite{hochreiter1997long}.
In this paper, we will focus on the ``addition'' and ``multiplication'' problems; due to space constraints, we refer the reader to \citep{hochreiter1997long} or \citep{sutskever2013importance} for their specification.
As proposed by \cite{hochreiter1997long}, we define accuracy as the proportion of sequences for which the absolute error between predicted value and the target value was less than .04.
Applying our feed-forward model to these tasks is somewhat disingenuous because they are commutative and therefore may be easier to solve with a model which ignores temporal order.
However, as we further argue in Section \ref{sec:discussion}, we believe these tasks provide a useful demonstration of our model's ability to refer to arbitrary locations in the input sequence when computing its output.

\subsection{Model Details}

For all experiments, we used the following model:
First, the state $h_t$ was computed from the input at each time step $x_t$ by $h_t = \textrm{LReLU}(W_{xh}x_t + b_{xh})$ where $W_{xh} \in \mathbb{R}^{D \times 2}, b_{xh} \in \mathbb{R}^D$ and $\textrm{LReLU}(x) = \max(x, .01x)$ is the ``leaky rectifier'' nonlinearity, as proposed by \cite{maas2013rectifier}.
We found that this nonlinearity improved early convergence so we used it in all of our models.
We tested models where the context vector $c$ was then computed either as in Equation (\ref{eq:ffattention}), with $a(h_t) =\tanh(W_{hc}h_t + b_{hc})$
where $W_{hc} \in \mathbb{R}^{1 \times D}, b_{hc} \in \mathbb{R}$, or simply as the unweighted mean of $h$ as in Equation (\ref{eq:unweighted}).
We then computed an intermediate vector $s = \textrm{LReLU}(W_{cs}c + b_{cs})$ where $W_{cs} \in \mathbb{R}^{D \times D}, b \in \mathbb{R}^D$ from which the output was computed as $y = \textrm{LReLU}(W_{sy}s + b_{sy})$ where $W_{sy} \in \mathbb{R}^{1 \times D}$, $b_{sy} \in \mathbb{R}$.
For all experiments, we set $D = 100$.

We used the squared error of the output $y$ against the target value for each sequence as an objective.
Parameters were optimized using ``adam'', a recently proposed stochastic optimization technique \citep{kingma2014adam}, with the optimization hyperparameters $\beta_1$ and $\beta_2$ set to the values suggested by \cite{kingma2014adam} (.9 and .999 respectively).
All weight matrices were initialized with entries drawn from a Gaussian distribution with a mean of zero and, for a matrix $W \in \mathbb{R}^{M \times N}$, a standard deviation of $1/\sqrt{N}$.
All bias vectors were initialized with zeros.
We trained on mini-batches of 100 sequences and computed the accuracy on a held-out test set of 1000 sequences every epoch, defined as 1000 parameter updates.
We stopped training when either 100\% accuracy was attained on the test set, or after 100 epochs.
All networks were implemented using Lasagne \citep{dieleman2015lasagne}, which is built on top of Theano \citep{bastien2012theano,bergstra2010theano}.

\begin{table}
  \centering
  \footnotesize
  \begin{tabular}{r c c c c c c c c c c c c}
    \toprule
    Task & \multicolumn{6}{c}{\textbf{Addition}} & \multicolumn{6}{c}{\textbf{Multiplication}} \\
    $T_0$ & 50 & 100 & 500 & 1000 & 5000 & 10000 & 50 & 100 & 500 & 1000 & 5000 & 10000 \\
    \cmidrule(r){2-7}
    \cmidrule(r){8-13}
    Attention & 1 & 1 & 1 & 1 & 2 & 3 & 1 & 2 & 4 & 2 & 15 & 6 \\
    Unweighted & 1 & 1 & 1 & 2 & 8 & 17 & 2 & 2 & 8 & 33 & \textcolor{gray}{89.8\%} & \textcolor{gray}{80.8\%} \\
    \bottomrule
  \end{tabular}
  \caption{Number of epochs required to achieve perfect accuracy, or accuracy after 100 epochs (greyed-out values), for the experiment described in Section \ref{sec:fixed}.}
  \label{tab:fixed}
\end{table}

\subsection{Fixed-Length Experiment}
\label{sec:fixed}

Traditionally, the sequence lengths tested in each task vary uniformly between $[T_0, 1.1T_0]$ for different values of $T_0$.
As $T_0$ increases, the model must be able to handle longer-term dependencies.
The largest value of $T_0$ attained using RNNs with different training, regularization, and model structures has varied from a few hundred \citep{martens2011learning,sutskever2013importance,le2015simple,krueger2015regularizing,arjovsky2015unitary} to a few thousand \citep{hochreiter1997long,jaegar2012long}.
We therefore tested our proposed feed-forward attention models for $T_0 \in \{50, 100, 500, 1000, 5000, 10000\}$.
The required number of epochs or accuracy after 100 epochs for each task, sequence length, and temporal integration method (adaptively weighted attention or unweighted mean) is shown in Table \ref{tab:fixed}.
For fair comparison, we report the best result achieved using any learning rate in $\{.0003, .001, .003, .01\}$.
From these results, it's clear that the feed-forward attention model can quickly solve these long-term memory problems for all sequence lengths we tested.
Our model is also efficient: Processing one epoch of 100,000 sequences with $T_0 = 10000$ took 254 seconds using an NVIDIA GTX 980 Ti GPU, while processing the same data with a single-layer vanilla RNN with a hidden dimensionality of 100 (resulting in a comparable number of parameters) took 917 seconds on the same hardware.
In addition, there is a clear benefit to using the attention mechanism of Equation (\ref{eq:ffattention}) instead of a simple unweighted average over time, which only incurs a marginal increase in the number of parameters (10,602 vs.\ 10,501, or less than 1\%).

\subsection{Variable-length Experiment}

Because the range of sequence lengths $[T_0, 1.1T_0]$ is small compared to the range of $T_0$ values we evaluated, we further tested whether it was possible to train a single model which could cope with sequences with highly varying lengths.
To our knowledge, such a variant of these tasks has not been studied before.
We trained models of the same architecture used in the previous experiment on minibatches of sequences whose lengths were chosen uniformly at random between 50 and 10000 time steps.
Using the attention mechanism of Equation (\ref{eq:ffattention}), on held-out test sets of 1000 sequences, our model achieved 99.9\% accuracy on the addition task and 99.4\% on the multiplication task after training for 100 epochs.
This suggests that a single feed-forward network with attention can simultaneously handle both short and very long sequences, with a marginal decrease in accuracy.
Using an unweighted average over time, we were only able to achieve accuracies of 77.4\% and 55.5\% on the variable-length addition and multiplication tasks, respectively.

\subsection{Discussion}
\label{sec:discussion}

A clear limitation of our proposed model is that it will fail on any task where temporal order matters because computing an average over time discards order information.
For example, on the two-symbol temporal order task \citep{hochreiter1997long} where a sequence must be classified in terms of whether two symbols $X$ and $Y$ appear in the order $X, X$; $Y, Y$; $X, Y$; or $Y, X$, our model can differentiate between the $X, X$ and $Y, Y$ cases perfectly but cannot differentiate between the $X, Y$ and $Y, X$ cases at all.
Nevertheless, we submit that for some real-world tasks involving sequential data, temporal order is substantially less important than being able to handle very long sequences.
For example, in Joachims' seminal paper on text document categorization \citep{joachims1998text}, he posits that ``word stems work well as representation units and that their ordering in a document is of minor importance for many tasks''.
In fact, the current state-of-the-art system for document classification still uses order-agnostic sequence integration \citep{lei2015molding}.
We have also shown in parallel work that our proposed feed-forward attention model can be used effectively for pruning large-scale (sub)sequence retrieval searches, even when the sequences are very long and high-dimensional \citep{raffel2016pruning}.

Our experiments explicitly demonstrate that including an attention mechanism can allow a model to refer to specific points in a sequence when computing its output.
They also provide an alternate argument for the claim made by \cite{bahdanau2014neural} that attention helps models handle very long and widely variable-length sequences.
We are optimistic that our proposed feed-forward model will prove beneficial in additional real-world problems requiring order-agnostic temporal integration of long sequences.
Further investigation is warranted; to facilitate future work, all of the code used in our experiments is available online.\footnote{\href{https://github.com/craffel/ff-attention/tree/master/toy_problems}{\texttt{https://github.com/craffel/ff-attention/tree/master/toy\char`_problems}}}

\section{Acknowledgements}

We thank Sander Dieleman, Bart van Merri{\"e}nboer, S{\o}ren Kaae S{\o}nderby, Brian McFee, and our anonymous reviewers for discussion and feedback.

\clearpage

\bibliography{refs}

\begin{thebibliography}{27}
\providecommand{\natexlab}[1]{#1}
\providecommand{\url}[1]{\texttt{#1}}
\expandafter\ifx\csname urlstyle\endcsname\relax
  \providecommand{\doi}[1]{doi: #1}\else
  \providecommand{\doi}{doi: \begingroup \urlstyle{rm}\Url}\fi

\bibitem[Arjovsky et~al.(2015)Arjovsky, Shah, and Bengio]{arjovsky2015unitary}
Martin Arjovsky, Amar Shah, and Yoshua Bengio.
\newblock Unitary evolution recurrent neural networks.
\newblock \emph{arXiv:1511.06464}, 2015.

\bibitem[Bahdanau et~al.(2014)Bahdanau, Cho, and Bengio]{bahdanau2014neural}
Dzmitry Bahdanau, Kyunghyun Cho, and Yoshua Bengio.
\newblock Neural machine translation by jointly learning to align and
  translate.
\newblock \emph{arXiv:1409.0473}, 2014.

\bibitem[Bahdanau et~al.(2015)Bahdanau, Chorowski, Serdyuk, Brakel, and
  Bengio]{bahdanau2015end}
Dzmitry Bahdanau, Jan Chorowski, Dmitriy Serdyuk, Philemon Brakel, and Yoshua
  Bengio.
\newblock End-to-end attention-based large vocabulary speech recognition.
\newblock \emph{arXiv:1508.04395}, 2015.

\bibitem[Bastien et~al.(2012)Bastien, Lamblin, Pascanu, Bergstra, Goodfellow,
  Bergeron, Bouchard, Warde-Farley, and Bengio]{bastien2012theano}
Fr{\'e}d{\'e}ric Bastien, Pascal Lamblin, Razvan Pascanu, James Bergstra, Ian
  Goodfellow, Arnaud Bergeron, Nicolas Bouchard, David Warde-Farley, and Yoshua
  Bengio.
\newblock Theano: new features and speed improvements.
\newblock In \emph{Deep Learning and Unsupervised Feature Learning NIPS 2012
  Workshop}, 2012.

\bibitem[Bengio et~al.(1994)Bengio, Simard, and Frasconi]{bengio1994learning}
Yoshua Bengio, Patrice Simard, and Paolo Frasconi.
\newblock Learning long-term dependencies with gradient descent is difficult.
\newblock \emph{IEEE Transactions on Neural Networks}, 5\penalty0 (2):\penalty0
  157--166, 1994.

\bibitem[Bergstra et~al.(2010)Bergstra, Breuleux, Bastien, Lamblin, Pascanu,
  Desjardins, Turian, Warde-Farley, and Bengio]{bergstra2010theano}
James Bergstra, Olivier Breuleux, Fr{\'e}d{\'e}ric Bastien, Pascal Lamblin,
  Razvan Pascanu, Guillaume Desjardins, Joseph Turian, David Warde-Farley, and
  Yoshua Bengio.
\newblock Theano: a {CPU} and {GPU} math expression compiler.
\newblock In \emph{Proceedings of the Python for scientific computing
  conference (SciPy)}, 2010.

\bibitem[Chan et~al.(2015)Chan, Jaitly, Le, and Vinyals]{chan2015listen}
William Chan, Navdeep Jaitly, Quoc~V. Le, and Oriol Vinyals.
\newblock Listen, attend and spell.
\newblock \emph{arXiv:1508.01211}, 2015.

\bibitem[Cho(2015)]{cho2015introduction}
Kyunghyun Cho.
\newblock Introduction to neural machine translation with {GPUs} (part 3).
\newblock
  \href{http://devblogs.nvidia.com/parallelforall/introduction-neural-machine-translation-gpus-part-3/}{\texttt{http://devblogs.nvidia.com/\allowbreak
  parallelforall/\allowbreak introduction-\allowbreak neural-\allowbreak
  machine-\allowbreak translation-\allowbreak gpus-\allowbreak part\allowbreak
  -3/}}, 2015.

\bibitem[Dieleman(2014)]{dieleman2014recommending}
Sander Dieleman.
\newblock Recommending music on {Spotify} with deep learning.
\newblock
  \href{http://benanne.github.io/2014/08/05/spotify-cnns.html}{\texttt{http://benanne.github.io/\allowbreak
  2014/\allowbreak 08/\allowbreak 05/\allowbreak spotify-cnns.html}}, 2014.

\bibitem[Dieleman et~al.(2015)Dieleman, Schl{\"u}ter, Raffel, Olson, and
  Sonderby]{dieleman2015lasagne}
Sander Dieleman, Jan Schl{\"u}ter, Colin Raffel, Eben Olson, and Soren~Kaae
  Sonderby.
\newblock Lasagne: First release.
\newblock \href{https://github.com/Lasagne/Lasagne}{\texttt{https://\allowbreak
  github.com/\allowbreak Lasagne/\allowbreak Lasagne}}, 2015.

\bibitem[Graves et~al.(2014)Graves, Wayne, and Danihelka]{graves2014neural}
Alex Graves, Greg Wayne, and Ivo Danihelka.
\newblock Neural turing machines.
\newblock \emph{arXiv:1410.5401}, 2014.

\bibitem[Hochreiter \& Schmidhuber(1997)Hochreiter and
  Schmidhuber]{hochreiter1997long}
Sepp Hochreiter and J{\"u}rgen Schmidhuber.
\newblock Long short-term memory.
\newblock \emph{Neural computation}, 9\penalty0 (8):\penalty0 1735--1780, 1997.

\bibitem[Jaeger(2012)]{jaegar2012long}
Herbert Jaeger.
\newblock Long short-term memory in echo state networks: Details of a
  simulation study.
\newblock Technical Report~27, Jacobs University, 2012.

\bibitem[Joachims(1998)]{joachims1998text}
Thorsten Joachims.
\newblock \emph{Text categorization with support vector machines: Learning with
  many relevant features}.
\newblock Springer, 1998.

\bibitem[Kingma \& Ba(2014)Kingma and Ba]{kingma2014adam}
Diederik Kingma and Jimmy Ba.
\newblock Adam: A method for stochastic optimization.
\newblock \emph{arXiv:1412.6980}, 2014.

\bibitem[Krueger \& Memisevic(2015)Krueger and
  Memisevic]{krueger2015regularizing}
David Krueger and Roland Memisevic.
\newblock Regularizing {RNNs} by stabilizing activations.
\newblock \emph{arXiv:1511.08400}, 2015.

\bibitem[Le et~al.(2015)Le, Jaitly, and Hinton]{le2015simple}
Quoc~V. Le, Navdeep Jaitly, and Geoffrey~E. Hinton.
\newblock A simple way to initialize recurrent networks of rectified linear
  units.
\newblock \emph{arXiv:1504.00941}, 2015.

\bibitem[Lei et~al.(2015)Lei, Barzilay, and Jaakkola]{lei2015molding}
Tao Lei, Regina Barzilay, and Tommi Jaakkola.
\newblock Molding {CNNs} for text: non-linear, non-consecutive convolutions.
\newblock In \emph{Proceedings of the 2015 Conference on Empirical Methods in
  Natural Language Processing}, pp.\  1565--1575, 2015.

\bibitem[Maas et~al.(2013)Maas, Hannun, and Ng]{maas2013rectifier}
Andrew~L. Maas, Awni~Y. Hannun, and Andrew~Y. Ng.
\newblock Rectifier nonlinearities improve neural network acoustic models.
\newblock In \emph{ICML Workshop on Deep Learning for Audio, Speech, and
  Language Processing}, 2013.

\bibitem[Martens \& Sutskever(2011)Martens and Sutskever]{martens2011learning}
James Martens and Ilya Sutskever.
\newblock Learning recurrent neural networks with hessian-free optimization.
\newblock In \emph{Proceedings of the 28th International Conference on Machine
  Learning}, pp.\  1033--1040, 2011.

\bibitem[Pascanu et~al.(2012)Pascanu, Mikolov, and
  Bengio]{pascanu2012difficulty}
Razvan Pascanu, Tomas Mikolov, and Yoshua Bengio.
\newblock On the difficulty of training recurrent neural networks.
\newblock \emph{arXiv:1211.5063}, 2012.

\bibitem[Raffel \& Ellis(2016)Raffel and Ellis]{raffel2016pruning}
Colin Raffel and Daniel P.~W. Ellis.
\newblock Pruning subsequence search with attention-based embedding.
\newblock In \emph{Proceedings of the 41st IEEE International Conference on
  Acoustics, Speech, and Signal Processing}, 2016.

\bibitem[S{\o}nderby et~al.(2015)S{\o}nderby, S{\o}nderby, Nielsen, and
  Winther]{sonderby2015convolutional}
S{\o}ren~Kaae S{\o}nderby, Casper~Kaae S{\o}nderby, Henrik Nielsen, and Ole
  Winther.
\newblock Convolutional lstm networks for subcellular localization of proteins.
\newblock \emph{arXiv:1503.01919}, 2015.

\bibitem[Sukhbaatar et~al.(2015)Sukhbaatar, Szlam, Weston, and
  Fergus]{sukhbaatar2015end}
Sainbayar Sukhbaatar, Arthur Szlam, Jason Weston, and Rob Fergus.
\newblock End-to-end memory networks.
\newblock \emph{arXiv:1503.08895}, 2015.

\bibitem[Sutskever et~al.(2013)Sutskever, Martens, Dahl, and
  Hinton]{sutskever2013importance}
Ilya Sutskever, James Martens, George Dahl, and Geoffrey Hinton.
\newblock On the importance of initialization and momentum in deep learning.
\newblock In \emph{Proceedings of the 30th International Conference on Machine
  Learning}, pp.\  1139--1147, 2013.

\bibitem[Werbos(1990)]{werbos1990backpropagation}
Paul~J. Werbos.
\newblock Backpropagation through time: what it does and how to do it.
\newblock \emph{Proceedings of the IEEE}, 78\penalty0 (10):\penalty0
  1550--1560, 1990.

\bibitem[Xu et~al.(2015)Xu, Ba, Kiros, Courville, Salakhutdinov, Zemel, and
  Bengio]{xu2015show}
Kelvin Xu, Jimmy Ba, Ryan Kiros, Aaron Courville, Ruslan Salakhutdinov, Richard
  Zemel, and Yoshua Bengio.
\newblock Show, attend and tell: Neural image caption generation with visual
  attention.
\newblock \emph{arXiv:1502.03044}, 2015.

\end{thebibliography}
\bibliographystyle{iclr2016_workshop}

\end{document}